%% file: 2026-kelly-sliding-mfi.tex
\documentclass[letterpaper,10pt,conference]{ieeeconf}
\IEEEoverridecommandlockouts %
\usepackage{xcolor}
\usepackage{graphicx}
\graphicspath{{figs/}}

\usepackage{amsmath}
\usepackage{amssymb}
\usepackage{amsfonts}
\usepackage{amstext}

\usepackage{amsthm}
\usepackage{bm}
\usepackage{mathtools}
\allowdisplaybreaks

\makeatletter
\renewcommand{\thetable}{\arabic{table}}
\def\fnum@table{Tab.\nobreakspace\thetable}
\def\fnum@figure{Fig.\nobreakspace\thefigure}
\long\def\@makecaption#1#2{%
  \vskip\abovecaptionskip
  \sbox\@tempboxa{\footnotesize #1.\ #2}%
  \ifdim \wd\@tempboxa >\hsize
    {\footnotesize #1.\ #2\par}%
  \else
    \global\@minipagefalse
    \hb@xt@\hsize{\hfil\box\@tempboxa\hfil}%
  \fi
  \vskip\belowcaptionskip}
\makeatother

\usepackage{booktabs}
\usepackage{multirow}

\theoremstyle{definition}

\usepackage[
  backend=biber,
  hyperref=true,
  sorting=none,
  style=ieee-caps]
{biblatex}

\usepackage[noabbrev]{cleveref}
\crefname{equation}{}{}
\crefname{table}{Tab.}{Tabs.}
\crefname{figure}{Fig.}{Figs.}

\input{notation-shortcuts}

\input{notation-specifics}

\title{\Large\bf A Sliding Window Filter on the Galilean Group for Consistent\\ Aided Inertial Navigation with Unknown Measurement Delays}

\author{Jonathan Kelly$^{1}$
\thanks{$^{1}$Space \& Terrestrial Autonomous Robotic Systems (STARS) Laboratory, University of Toronto Institute for Aerospace Studies (UTIAS), Toronto, Canada. \tt\footnotesize jonathan.kelly@robotics.utias.utoronto.ca}
\thanks{This research was supported in part by the Canada Research Chairs program. Jonathan Kelly is a Vector Institute Faculty Affiliate.}}

\begin{document}
\maketitle

\begin{abstract}
We study aided inertial navigation when the aiding sensor measurements are subject to an unknown constant delay.
The goal is to estimate the delay and navigation state jointly so that delayed measurements correct the trajectory at the appropriate times, yielding a more accurate navigation solution.
We formulate the problem on the special Galilean group, which provides a natural state-space structure for aided navigation with uncertainty in both motion and timing.
We then examine the observability of joint delay and state estimation and show that, for a single delayed measurement, the model admits an exact symmetry in which a change in the delay can be compensated by a change in the navigation state, leaving the measurement unchanged.
Processing measurements individually allows spurious information to `leak' along the corresponding null direction of the measurement Jacobian, producing overconfident and inconsistent estimates.
Applying measurements from multiple times together can eliminate this direction when the trajectory is informative enough.
Motivated by this result, we develop a sliding window filter that retains a short history of navigation states and applies delayed aiding corrections jointly across the active window.
We conduct a series of simulation studies to characterize estimator accuracy and consistency.
The simulations demonstrate that an estimator that does not maintain an adequate window can rapidly become highly inconsistent, whereas even a short sliding window markedly improves consistency by providing the temporal support that observability requires.
\end{abstract}

\section{Introduction}
\label{sec:introduction}

Aided inertial navigation systems provide real-time estimates of orientation, velocity, and position for guidance, control, and autonomy in a wide range of aerospace and robotics applications.
High-rate angular-velocity and specific-force measurements from an inertial measurement unit (IMU) are integrated to propagate the navigation solution; a Bayesian estimator, typically an extended Kalman filter, fuses the inertial data with lower-rate aiding measurements to correct drift and maintain long-term accuracy~\cite{2008_Farrell_Aided}.

In practice, measurements from different devices are rarely perfectly synchronized, particularly in low-cost systems assembled from discrete sensing and computing components.
Independent clocks, acquisition and processing delays, buffering, and communication latency can introduce relative timing offsets between the aiding measurements and the IMU data stream.
Such delays are often modelled as constant but unknown~\cite{2014_Kelly_Determining,2014_Li_Online}.
When the delay cannot be measured or calibrated independently, it must be estimated jointly with the navigation state.

Several online methods handle an unknown measurement delay by augmenting the estimator state with the delay parameter~\cite{2011_Skog_Time,2014_Li_Online}.
This approach is straightforward and leaves the basic filtering architecture largely unchanged.
However, a delayed aiding measurement constrains the trajectory at an earlier time, coupling the delay to the navigation state and vehicle motion.
Our prior work showed that such augmented-state filters have fundamental structural limitations that can lead to inconsistency and possible divergence~\cite{2021_Kelly_Question}.

In this paper, we examine the observability of joint delay and navigation-state estimation.
For a single delayed measurement, the model admits a symmetry: a change in the delay can be compensated by a change in the navigation state, leaving the measurement unchanged.
Because the compensation is exact, the symmetry is present under any additive, invariant, or equivariant error formulation, and appears in the linearized model as a delay--state null direction.
Processing measurements individually allows spurious information to `leak' along this unobservable direction, leading to inconsistency~\cite{2017_Huang_Towards,2023_Lisus_Know}.
Preventing information leak requires sufficiently independent constraints across time.

Motivated by this observability result, we develop a sliding window filter on the special Galilean group, $\LieGroupSGal{3}$.
The group provides a natural state-space structure for inertial navigation by incorporating orientation, velocity, position, and time within a single group element.
The filter retains a short history of Galilean transformations, interpolates on $\LieGroupSGal{3}$ to predict measurements at arbitrary delayed times, and applies measurements jointly across the window to constrain the delay and navigation states.
We characterize the proposed filtering architecture through simulations, comparing it with a buffered extended Kalman filter (EKF) that stores a portion of the integrated IMU history to reconstruct past transformations for delayed updates, similar to~\cite{2026_Delama_Galilean}.
We make the following contributions.
\begin{itemize}
\item We develop a sliding window filter on $\LieGroupSGal{3}$ that uses measurements from different times to jointly constrain the delay and navigation states.
\item We introduce an interpolation-based measurement update on $\LieGroupSGal{3}$, allowing predicted measurements to be computed at arbitrary times within the active window.
\item We show that observability of the delay and navigation state depends on both the trajectory and the aiding measurement history.
\item We demonstrate through simulation that the buffered EKF rapidly becomes inconsistent, whereas even a short sliding window markedly improves estimation accuracy and consistency.
\end{itemize}

The remainder of the paper is organized as follows.
\Cref{sec:related_work} reviews related work, while \Cref{sec:preliminaries} provides the required background material. %
\Cref{sec:aided_navigation} develops our sliding window filter, \Cref{sec:observability} analyzes observability, and \Cref{sec:simulation_studies} presents our simulation studies.
\Cref{sec:conclusion} concludes and discusses several directions for future work.

\section{Related Work}
\label{sec:related_work}

Extended or error-state Kalman filters and fixed-lag smoothers are the usual choice for aided inertial navigation, typically implemented in a predictor-corrector form~\cite{2008_Farrell_Aided}.
Online temporal calibration has been studied in visual-inertial and multi-sensor fusion, often by augmenting the state with a delay parameter~\cite{2010_Nilsson_Joint,2014_Li_Online,2019_Yang_Degenerate,2023_Yang_Online}.
The delay must, however, be observable from the trajectory and measurement history for the estimation problem to be well posed.

Lie-group methods are now standard for geometric state estimation in robotics~\cite{2024_Barfoot_State}, with $\LieGroupSE{3}$ and $\LieGroupSETwo{3}$ widely used for inertial navigation.
The special Galilean group $\LieGroupSGal{3}$ extends $\LieGroupSETwo{3}$, and hence $\LieGroupSE{3}$, to include time.
In robotics, $\LieGroupSGal{2}$ has been used for uncertainty representation~\cite{2021_Giefer_Uncertainties} and $\LieGroupSGal{3}$ for IMU preintegration~\cite{2019_Fourmy_Absolute,2024_Shalaby_Multi-Robot}.
Most recently, Delama et al.~\cite{2026_Delama_Galilean} developed an equivariant filter on $\LieGroupSGal{3}$ for joint navigation and delay estimation, storing IMU data to reconstruct delayed states while maintaining a single navigation-state estimate.
Our comparison filter uses a similar architecture, but without the equivariant error formulation; we refer to it as a \emph{buffered} EKF.%
\footnote{The term is ours: the estimator reconstructs past states from stored IMU data rather than retaining them explicitly.}

Sliding window filters and fixed-lag smoothers retain a bounded history of past states without solving a full batch problem~\cite{2010_Sibley_Sliding}.
Consistency in filtering with unobservable states has been studied extensively~\cite{2017_Huang_Towards}, and Lisus et al.~\cite{2023_Lisus_Know} show specifically that sliding window linearization and marginalization introduce spurious information when the unobservable directions are not preserved.
Our use of a window is motivated by these consistency concerns: retaining multiple past states allows delayed measurements to provide the cross-time constraints needed to prevent spurious information gain along the delay--state null direction.

\section{Preliminaries}
\label{sec:preliminaries}

In this section, we briefly describe the special Galilean group $\LieGroupSGal{3}$. %
We then discuss the notion of local observability used in \Cref{sec:observability}.

\subsection{The Special Galilean Group}
\label{subsec:SGal3_group}

The special Galilean group $\LieGroupSGal{3}$ is a $10$-dimensional Lie group that describes transformations between inertial reference frames in relative motion~\cite{2023_Kelly_Galilean,2025_Mahony_Galilean}.
In matrix form,
\begin{equation*}
\LieGroupSGal{3} \Defined
\left\{
\Matrix{X} =
\bbm
\Matrix{C} & \Vector{v} & \Vector{r} \\
\Zero & 1 & \eta \\
\Zero & 0 & 1
\ebm
\in \Real^{5 \times 5}
\ \middle\vert \
\Matrix{C} \in \LieGroupSO{3}
\right\},
\end{equation*}
where $\Matrix{C}$ is a rotation matrix, $\Vector{v} \in \Real^3$ is a velocity boost, $\Vector{r} \in \Real^3$ is a spatial translation, and $\eta \in \Real$ is a time translation.
The group operation is matrix multiplication. A group element acts on spacetime coordinates $(\Vector{p}, t)$, with $\Vector{p} \in \Real^3$, as $(\Vector{p},t) \mapsto (\Matrix{C}\<\Vector{p} + \Vector{v}\<t + \Vector{r},\, t + \eta)$.
The essential point is that time is part of the group, so composition couples time translation to spatial translation through the boost.

The Lie algebra $\LieAlgebraSGal{3}$ is the tangent space of $\LieGroupSGal{3}$ at the identity, with the matrix commutator as its Lie bracket.
Applied to $\Vector{\phi} \in \Real^3$, the operator $(\cdot)^{\wedge}$ produces a skew-symmetric matrix, giving the standard isomorphism between $\Real^3$ and $\LieAlgebraSO{3}$.
The same notation is overloaded for $\LieAlgebraSGal{3}$ via the map $(\cdot)^{\wedge} : \Real^{10} \to \LieAlgebraSGal{3}$ defined as
\begin{equation*}
\Vector{\xi}^{\wedge}
=
\bbm
\Vector{\rho} \\
\Vector{\nu} \\
\Vector{\phi} \\
\iota
\ebm^{\wedge}
=
\bbm
\Vector{\phi}^{\wedge} & \Vector{\nu} & \Vector{\rho} \\
\Zero & 0 & \iota \\
\Zero & 0 & 0
\ebm,
\end{equation*}
where $\Vector{\rho}$ parameterizes the spatial translation, $\Vector{\nu}$ the boost, $\Vector{\phi}$ the rotational part, and $\iota$ the time translation.
The corresponding operator $(\cdot)^{\vee}:\LieAlgebraSGal{3}\to\Real^{10}$ is the inverse of $(\cdot)^{\wedge}$, so that $\left(\Vector{\xi}^{\wedge}\right)^{\vee} = \Vector{\xi}$.
Closed-form expressions for the exponential and logarithmic maps of $\LieGroupSGal{3}$, written $\Matexp{\cdot}$ and $\Matlog{\cdot}$, are given in~\cite{2023_Kelly_Galilean}.

\subsection{Observability and Identifiability}
\label{subsec:observability}

We consider whether the delay and navigation state can be determined from the available measurements, adopting a local differential viewpoint consistent with classical nonlinear observability analysis~\cite{1977_Hermann_Nonlinear}.
A defining characteristic of the delayed-measurement problem is that observability cannot be determined at a single instant, since each aiding measurement constrains the state at an earlier, delay-dependent time.
The appropriate rank test therefore requires measurements over an interval rather than at a single point.
Full column rank of the stacked measurement Jacobian implies that the delay and state are locally distinguishable to first order, while a rank deficiency gives a null direction along which the state and the delay can be perturbed together with no effect on the residuals to first order.

Strictly, the delay is a constant parameter and not a state, so determining it is a question of local identifiability.
Because the navigation state is propagated forward from the inertial inputs, however, the delay acts as a shift along the trajectory.
The same rank test therefore applies to both questions, so we use the term observability throughout.

\section{Aided Navigation on the Galilean Group}
\label{sec:aided_navigation}

We now introduce our sliding window filter on $\LieGroupSGal{3}$ for aided navigation.
The process model is driven by IMU angular-rate and specific-force measurements, both resolved in the body frame.
The filter maintains an \emph{active window} of Galilean transformations together with the delay and the gyroscope and accelerometer biases.
Each transformation relates the body frame at the associated IMU clock time to the global inertial reference frame at the initial time.
We assume a known gravity vector and neglect the rotation of the Earth.

The active window serves two roles: it provides a local trajectory representation that can be adjusted by measurement updates, and it preserves the temporal structure needed for delay estimation.
Delayed aiding measurements are handled by interpolation on $\LieGroupSGal{3}$, allowing the trajectory to be queried at an arbitrary time as the delay estimate changes, without the need for repeated IMU integration.
As the window advances, older transformations are marginalized to maintain a bounded window size.

\subsection{State Representation}
\label{subsec:state_representation}

To keep the notation light, we omit explicit frame identifiers.
At update index $k$, the active window contains $n$ Galilean transformations, and the state is%
\footnote{The index $k$ advances each time a transformation is added to the window. Fewer than $n$ transformations are available immediately after initialization; the expressions here assume that the window has size $n$.}
\begin{equation}
\label{eqn:window_states}
\State_{k}
=
\left(
\Matrix{X}_{k-m},\, \ldots,\, \Matrix{X}_{k},\,
\tau,\,
\Vector{b}_{\omega},\,
\Vector{b}_{\rms{a}}
\right),
\end{equation}
where $m = n - 1$ and each transformation
\begin{equation}
\label{eqn:galilean_element}
\Matrix{X}_i
=
\bbm
\Matrix{C}_i & \Vector{v}_i & \Vector{r}_i \\
\Zero & 1 & t_i \\
\Zero & 0 & 1
\ebm
\in \LieGroupSGal{3},
\quad
i = k - m, \ldots, k
\end{equation}
collects the orientation $\Matrix{C}_i \in \LieGroupSO{3}$, velocity $\Vector{v}_i \in \Real^3$, and position $\Vector{r}_i \in \Real^3$ of the body at time $t_i$.
The delay $\tau$, the gyroscope bias $\Vector{b}_{\omega} \in \Real^3$, and the
accelerometer bias $\Vector{b}_{\rms{a}} \in \Real^3$ are shared across the
window, with the biases taken at $t_k$.

The uncertainty in $\State_k$ is represented locally by a multivariate Gaussian prior with covariance matrix $\Matrix{P}_k$, defined over the perturbation coordinates
\begin{equation}
\label{eqn:window_error_state}
\delta\SmallState_k
=
\bbm
\delta\Vector{\zeta}_{k-m}^{\T} &
\cdots &
\delta\Vector{\zeta}_{k}^{\T} &
\delta\tau &
\delta\Vector{b}_{\omega}^{\T} &
\delta\Vector{b}_{\rms{a}}^{\T}
\ebm^{\T}.
\end{equation}
Each Galilean transformation is perturbed on the right,
\begin{equation}
\label{eqn:perturbation}
\Matrix{X}_i
=
\bar{\Matrix{X}}_i
\Matexp{\bbm \delta\Vector{\zeta}_i \\ 0 \ebm^{\wedge}},
\quad\!
\delta\Vector{\zeta}_i
=
\bbm
\delta\Vector{\rho}_i^{\T}\!\! &
\delta\Vector{\nu}_i^{\T}\!\! &
\delta\Vector{\phi}_i^{\T}
\ebm^{\T}
\in \Real^9,
\end{equation}
where $\bar{(\cdot)}$ denotes a nominal value.
We treat the IMU clock as exact, so only the nine navigation degrees of freedom of each transformation have associated uncertainty, and $\Matrix{P}_k$ has size $(9n + 7) \times (9n + 7)$.

\subsection{Process Model}
\label{subsec:process_model}

The process model specifies how the state and the Gaussian prior are propagated from time $t_k$ to time $t_{k+1}$ while adding one transformation to the active window.
A copy of the most recent transformation $\Matrix{X}_k$ is first appended to the active window, and the prior covariance is augmented by stochastic cloning~\cite{2002_Roumeliotis_Stochastic}, so that the uncertainty of the copy is initially perfectly correlated with that of $\Matrix{X}_k$.
Propagating the copy with the IMU measurements over $[t_k,\,t_{k+1}]$ gives $\Matrix{X}_{k+1}$ and the corresponding propagated prior covariance.

Let $\Vector{\omega}_{\rms{m}}(t)$ and $\Vector{a}_{\rms{m}}(t)$ denote the measured angular rate and specific force, both resolved in the body frame.
The true angular rate and specific force are related to the IMU measurements by
\begin{equation*}
\begin{aligned}
\Vector{\omega}(t)
& =
\Vector{\omega}_{\rms{m}}(t)
-
\Vector{b}_{\omega}(t)
-
\Vector{n}_{\omega}(t),
\\
\Vector{s}(t)
& =
\Vector{a}_{\rms{m}}(t)
-
\Vector{b}_{\rms{a}}(t)
-
\Vector{n}_{\rms{a}}(t),
\end{aligned}
\end{equation*}
where $\Vector{n}_{\omega}(t)$ and $\Vector{n}_{\rms{a}}(t)$ are zero-mean white Gaussian noise processes with continuous-time spectral densities $\Matrix{Q}_{\omega}$ and $\Matrix{Q}_{\rms{a}}$, respectively.
The biases follow random walks,
\begin{equation*}
\dot{\Vector{b}}_{\omega}(t)
=
\Vector{n}_{\rms{b}\<\omega}(t),
\quad
\dot{\Vector{b}}_{\rms{a}}(t)
=
\Vector{n}_{\rms{ba}}(t),
\end{equation*}
driven by zero-mean white Gaussian noise processes with spectral densities $\Matrix{Q}_{\rms{b}\<\omega}$ and $\Matrix{Q}_{\rms{ba}}$, respectively, and all noises are mutually independent.
The delay is treated as an unknown constant, $\dot{\tau}=0$.
The continuous-time kinematics are
\begin{equation}
\label{eqn:component_kinematics}
\dot{\Matrix{C}}(t)
=
\Matrix{C}(t)\<\Vector{\omega}(t)^{\wedge},
\quad\!
\dot{\Vector{v}}(t)
=
\Matrix{C}(t)\<\Vector{s}(t)
+
\Vector{g},
\quad\!
\dot{\Vector{r}}(t)
=
\Vector{v}(t),
\end{equation}
where $\Vector{g}$ is the known gravity vector.
The time coordinate advances with the IMU clock, $\dot{t}=1$.

Since the time coordinates of two Galilean transformations sum under composition, the transformation at any time $t$ factors uniquely as
$\Matrix{X}(t)=\Matrix{T}(t)\<\Matrix{N}(t)$, where
\begin{equation*}
\Matrix{T}(\beta)
\Defined
\bbm
\Identity_3 & \Zero & \Zero \\
\Zero & 1 & \beta \\
\Zero & 0 & 1
\ebm,
\quad
\Matrix{N}(t)
\Defined
\bbm
\Matrix{C}(t) & \Vector{v}(t) & \Vector{r}(t) \\
\Zero & 1 & 0 \\
\Zero & 0 & 1
\ebm.
\end{equation*}
The factor $\Matrix{T}(\beta)$ is a time translation by $\beta$, while $\Matrix{N}(t)$ has zero time coordinate and so is a member of the \emph{isochronous} Galilean subgroup~\cite{2023_Kelly_Galilean}.
We write $\Matrix{N}_k\Defined\Matrix{N}(t_k)$.

The continuous-time kinematics of the isochronous factor are
\begin{equation}
\label{eqn:isochronous_kinematics}
\dot{\Matrix{N}}(t)
=
\Matrix{\Gamma}\<\Matrix{N}(t)
+
\Matrix{N}(t)\<\Matrix{\Omega}(t),
\end{equation}
where
\begin{equation*}
\Matrix{\Gamma}
\Defined
\bbm
\Zero & \Vector{g} & \Zero \\
\Zero & 0 & -1 \\
\Zero & 0 & 0
\ebm,
\quad
\Matrix{\Omega}(t)
\Defined
\bbm
\Vector{\omega}(t)^{\wedge} & \Vector{s}(t) & \Zero \\
\Zero & 0 & 1 \\
\Zero & 0 & 0
\ebm.
\end{equation*}
Expanding the products reproduces \Cref{eqn:component_kinematics} exactly.
For given bias-corrected inputs, the kinematics in \Cref{eqn:isochronous_kinematics} are group affine~\cite{2017_Barrau_Invariant}.
Let $\Matrix{\Phi}_k(\Delta t)\in\LieGroupSGal{3}$ denote the body-frame motion over the signed interval $\Delta t$ from $t_k$,
\begin{equation}
\label{eqn:phi_flow}
\frac{d}{d\<\Delta t}\<\Matrix{\Phi}_k(\Delta t)
=
\Matrix{\Phi}_k(\Delta t)\<\Matrix{\Omega}(t_k+\Delta t),
\quad
\Matrix{\Phi}_k(0)
=
\Identity_5,
\end{equation}
where the subscript marks the starting time and the argument the elapsed interval, so that $\Matrix{\Phi}_k(\Delta t)$ has time coordinate $\Delta t$.
Since $\Matrix{\Gamma}$ is constant, integration over
$\Delta t_k\Defined t_{k+1}-t_k$ gives
\begin{equation*}
\Matrix{N}_{k+1}
=
\Matrix{U}(\Delta t_k)\<
\Matrix{N}_k\<
\Matrix{\Phi}_k(\Delta t_k),
\end{equation*}
where
\begin{equation*}
\Matrix{U}(\beta)
\Defined
\Matexp{\beta\<\Matrix{\Gamma}}
=
\bbm
\Identity_3
&
\beta\<\Vector{g}
&
-\tfrac{1}{2}\<\beta^2\<\Vector{g}
\\
\Zero & 1 & -\beta \\
\Zero & 0 & 1
\ebm.
\end{equation*}
The gravity contribution acts on the left, in the inertial frame, while the integrated IMU inputs act on the right, in the body frame.
The time coordinates of the three factors sum to zero, so $\Matrix{N}_{k + 1}$ is isochronous, and the full transformation is $\Matrix{X}_{k + 1} = \Matrix{T}(t_{k + 1})\<\Matrix{N}_{k + 1}$.
The absolute clock time $t_{k + 1}$ appears only in the leading time translation, while the propagation depends on the elapsed interval and the corresponding IMU inputs.
The factored form is also the basis for the observability analysis in \Cref{sec:observability}.

The IMU noises perturb $\Matrix{\Phi}_{k}(\Delta t_{k})$, the bias random walks perturb the biases, and the existing state uncertainty is propagated through the process model.
Together, these determine the augmented prior covariance $\Matrix{P}_{k+1}$.
Only the covariance of $\Matrix{X}_{k+1}$, its cross-covariances with the other state variables, and the bias covariance are modified.
Aiding measurements are incorporated separately in the window update and do not modify the prior covariance while they remain active.
An equivariant preintegration scheme could also be used~\cite{2025_Delama_Equivariant}.

\subsection{Measurement Model}
\label{subsec:measurement_model}

Let an aiding measurement have arrival timestamp $t_{\rms{y}}$, providing an observation (possibly partial) of the navigation state at the delayed time $t_{\rms{d}} = t_{\rms{y}} - \tau$.
Associated with the measurement is the Galilean element
\begin{equation}
\label{eqn:measurement_element}
\Matrix{Y}
\Defined
\Matrix{T}(\tau)\<\Matrix{X}(t_{\rms{d}})
=
\Matrix{T}(t_{\rms{y}})\<\Matrix{N}(t_{\rms{d}}).
\end{equation}
Left multiplication by $\Matrix{T}(\tau)$ leaves the navigation components unchanged but advances the time coordinate from $t_{\rms{d}}$ to $t_{\rms{y}}$; the aiding sensor observes selected components of $\Matrix{N}(t_{\rms{d}})$.

At the current linearization point, the estimated delayed time is
$\bar{t}_{\rms{d}}=t_{\rms{y}}-\bar{\tau}$.
Let $\bar{\Matrix{X}}_j$ and $\bar{\Matrix{X}}_{j + 1}$ be the two transformations in the active window that bracket $\bar{t}_{\rms{d}}$, and define
\begin{equation*}
\bar{\kappa}
=
\frac{\bar{t}_{\rms{d}} - t_j}{\Delta t_j},
\quad
\Delta t_j
=
t_{j + 1} - t_j .
\end{equation*}
The delayed transformation is predicted by interpolating along the exponential curve connecting $\bar{\Matrix{X}}_j$ and $\bar{\Matrix{X}}_{j + 1}$,
\begin{equation*}
\bar{\Matrix{X}}_{\rms{d}}
=
\bar{\Matrix{X}}_j
\Matexp{\bar{\kappa}\,\Vector{\xi}_{j,j + 1}^{\wedge}},
\quad\!\!
\Vector{\xi}_{j,j + 1}
=
\Matlog{\Inv{\bar{\Matrix{X}}_j}\<\bar{\Matrix{X}}_{j + 1}}^{\vee}
\!\in \Real^{10},
\end{equation*}
and the predicted measurement is
\begin{equation*}
\bar{\Matrix{Y}}
=
\Matrix{T}(\bar{\tau})\<\bar{\Matrix{X}}_{\rms{d}}.
\end{equation*}
The time component of $\Vector{\xi}_{j,j + 1}$ is $\Delta t_j$, and so the time coordinate of $\bar{\Matrix{X}}_{\rms{d}}$ is
$t_j + \bar{\kappa}\<\Delta t_j = \bar{t}_{\rms{d}}$, and the time coordinate of $\bar{\Matrix{Y}}$ is $t_{\rms{y}}$, matching \Cref{eqn:measurement_element}.
As the delay estimate changes, the indices $j$ and $j + 1$ and the interpolation coefficient $\bar{\kappa}$ are recomputed, so the measurement support may shift within the active window.

We allow the effective delay of each aiding measurement to include independent timing jitter $n_{\rms{t}} \sim \mathcal{N}(0,\sigma_{\rms{t}}^2)$.
The navigation state associated with the measurement is therefore evaluated at
$t_{\rms{y}} - \tau - n_{\rms{t}}$.
Incorporating measurement noise
$\Vector{n}_{\rms{s}} \in \Real^9$, with covariance
$\Matrix{R}_{\rms{s}}$, the noisy measurement element is
\begin{equation}
\label{eqn:noisy_measurement}
\Perturbed{\Matrix{Y}}
=
\Matrix{T}(\tau + n_{\rms{t}})\<
\Matrix{X}(t_{\rms{y}} - \tau - n_{\rms{t}})\<
\Matexp{\bbm \Vector{n}_{\rms{s}} \\ 0 \ebm^{\wedge}}.
\end{equation}
Using $\Matrix{X}(t)=\Matrix{T}(t)\<\Matrix{N}(t)$,
\begin{equation*}
\Matrix{T}(\tau + n_{\rms{t}})\<
\Matrix{X}(t_{\rms{y}} - \tau - n_{\rms{t}})
=
\Matrix{T}(t_{\rms{y}})\<
\Matrix{N}(t_{\rms{y}}-\tau - n_{\rms{t}}).
\end{equation*}
Jitter does not change the time coordinate of the measurement element; instead, it changes the time at which the trajectory is queried.%
\footnote{Unlike a model with a noisy \emph{reported} timestamp, we treat the reported arrival time as known and model jitter in the latency between acquisition and arrival. The jitter therefore shifts the effective acquisition time along the trajectory.}

Writing $\tau=\bar{\tau}+\delta\tau$, the time at which the measurement is evaluated becomes $t_{\rms{y}}-\tau-n_{\rms{t}} = \bar{t}_{\rms{d}} - \left(\delta\tau+n_{\rms{t}}\right)$.
Since the times $t_j$ and $t_{j + 1}$ are fixed, the corresponding perturbation of the interpolation coefficient is
\begin{equation*}
\delta\kappa
=
-\frac{\delta\tau + n_{\rms{t}}}{\Delta t_j}.
\end{equation*}
Linearizing the noisy measurement model in \Cref{eqn:noisy_measurement} about the nominal prediction
$\bar{\Matrix{Y}}=\Matrix{T}(\bar{\tau})\<\bar{\Matrix{X}}_{\rms{d}}$
gives the measurement residual in right-perturbation coordinates
\begin{equation}
\label{eqn:linearized_residual}
\Vector{\epsilon}
=
\Matrix{S}
\left(
\Matrix{A}_j\<\delta\Vector{\zeta}_j
+
\Matrix{A}_{j + 1}\<\delta\Vector{\zeta}_{j + 1}
+
\left(\delta\tau + n_{\rms{t}}\right)\<\Vector{d}_{\tau}
+
\Vector{n}_{\rms{s}}
\right),
\end{equation}
where $\Matrix{S}$ is a selection matrix that extracts the observed navigation components, and $\Matrix{A}_j$ and $\Matrix{A}_{j + 1}$ are the interpolation Jacobians with respect to the two bracketing transformations.
The delay sensitivity $\Vector{d}_{\tau}\in\Real^9$ combines the effects of the leading time translation and the perturbation of the interpolation point,
\begin{equation}
\label{eqn:delay_sensitivity}
\bbm
\Vector{d}_{\tau}^{\T} & 0
\ebm^{\T}
=
\bbm
-\bar{\Vector{v}}_{\rms{d}}^{\T}\<\bar{\Matrix{C}}_{\rms{d}}
&
\Zero^{\T}
&
\Zero^{\T}
&
1
\ebm^{\T}
-
\frac{\Vector{\xi}_{j,j + 1}}{\Delta t_j},
\end{equation}
where $\bar{\Vector{v}}_{\rms{d}}$ and $\bar{\Matrix{C}}_{\rms{d}}$ are the velocity and orientation components of $\bar{\Matrix{X}}_{\rms{d}}$.
Since $\Vector{\xi}_{j,j + 1}$ has the time component $\Delta t_j$, the time components in \Cref{eqn:delay_sensitivity} cancel.

For full-state, pose, and position-only measurements, $\Matrix{S}$ selects nine, six, and three navigation components, respectively.
Because $n_{\rms{t}}$ and $\Vector{n}_{\rms{s}}$ are independent, marginalizing out the timing jitter gives the effective measurement covariance
\begin{equation*}
\Matrix{R}_{\rms{d}}
=
\Matrix{S}
\left(
\Matrix{R}_{\rms{s}} + \sigma_{\rms{t}}^2\, \Vector{d}_{\tau}\<\Vector{d}_{\tau}^{\T}
\right)
\Matrix{S}^{\T}.
\end{equation*}
To first order, the jitter changes the measurement covariance but not the deterministic measurement Jacobian used in the observability analysis of \Cref{sec:observability}.
Within each Gauss--Newton update, $\Matrix{R}_{\rms{d}}$ is evaluated at the first linearization of the measurement and held fixed throughout subsequent iterations until convergence.

\subsection{Gauss--Newton Update and Marginalization}
\label{subsec:filter_update}

At any time, the filter maintains an active set $\mathcal{M}_k$ of aiding measurements whose estimated delayed times lie within the active window.
New measurements are added as they arrive; once a measurement leaves the window, it is retired permanently.
All measurements in $\mathcal{M}_k$ are processed jointly.

Let $\Vector{q}(\State_k)$ denote the deviation from the prior mean, expressed in the local coordinates of \Cref{eqn:window_error_state}, and let $\Vector{e}_{\ell}(\State_k)$ denote the nonlinear residual for measurement $\ell$, whose first-order linearization is given by \Cref{eqn:linearized_residual}.
The maximum \mbox{a posteriori} estimate minimizes
\begin{equation}
\label{eqn:window_map_cost}
\begin{aligned}
J(\State_k)
={} &
\frac{1}{2}
\Vector{q}(\State_k)^{\T}\<
\Inv{\Matrix{P}_k}\<
\Vector{q}(\State_k)
\\
& +
\frac{1}{2}
\sum_{\ell\<\in \mathcal{M}_k}
\Vector{e}_{\ell}(\State_k)^{\T}\<
\Inv{\Matrix{R}_{\rms{d},\ell}}\<
\Vector{e}_{\ell}(\State_k).
\end{aligned}
\end{equation}
We solve \Cref{eqn:window_map_cost} by Gauss--Newton, applying transformation increments on the right as in \Cref{eqn:perturbation} and updating the delay and biases additively.
At each iteration, the delayed times, bracketing indices, interpolation coefficients, residuals, and Jacobians are recomputed.
The prior covariance $\Matrix{P}_k$ remains fixed throughout these iterations and when additional measurements are added to $\mathcal{M}_k$.

We use a covariance-form implementation, in which the IMU propagation is captured entirely by $\Matrix{P}_k$, so no separate process terms appear in \Cref{eqn:window_map_cost}.
If required, a local posterior covariance may be obtained after convergence from the inverse Gauss--Newton information matrix, but this posterior is not carried forward as the prior while the same measurements remain active.

When the window advances, only the measurements that depend on the departing transformation are incorporated into the Gaussian prior, using their linearizations at the converged full-window estimate.
The departing transformation is then marginalized from the resulting Gaussian, producing the prior for the reduced window.
In covariance form, this amounts to retaining the mean and covariance blocks associated with the remaining variables.
This procedure preserves information from retired measurements without double counting the measurements that remain active.
Each retired measurement is linearized once, at an estimate in which the delay--state direction is already constrained.

\section{Observability Under Measurement Delays}
\label{sec:observability}

In this section, we examine the joint observability of the delay, navigation state, and biases under delayed aiding measurements.
Throughout, we analyze the underlying estimation problem in its minimal parameterization: a single navigation state with nine coordinates, the delay, and the six biases.
We show that, for any single aiding measurement, the model admits an exact symmetry in which a change in the delay can be compensated by a change in the navigation state, leaving the measurement unchanged.
The symmetry defines a one-dimensional family of equivalent delay--state pairs; we derive it directly and then as a null direction of the measurement Jacobian.
The practical concern is that a single-measurement update can still reduce the estimator covariance along the null direction, contributing to the inconsistency demonstrated in \Cref{sec:simulation_studies}.
Stacking measurements in a windowed update can break the symmetry, provided that the trajectory is sufficiently informative.
We establish necessary measurement-count conditions for observability.

\subsection{Single Measurements and Symmetry}
\label{subsec:measurement_symmetry}

Consider an aiding measurement of the full navigation state that arrives with timestamp $t_{\rms{y}}$ and corresponds to the state at the earlier delayed time $t_{\rms{d}} = t_{\rms{y}} - \tau$.
Without loss of generality, we take $\Matrix{N}_k$ as the reference, so the trajectory is
\begin{equation}
\label{eqn:local_trajectory}
\Matrix{N}(t_k + \Delta t)
=
\Matrix{U}(\Delta t)\<
\Matrix{N}_k\<
\Matrix{\Phi}_k(\Delta t),
\end{equation}
where $\Matrix{\Phi}_k$ is given by \Cref{eqn:phi_flow}.
The measurement has time offset $\Delta t_{\rms{d}} = t_{\rms{d}} - t_k$, which is negative since $t_{\rms{d}}$ precedes $t_k$.

Suppose the delay is $\tau + \Delta\tau$ rather than $\tau$.
The measurement is then at offset $\Delta t_{\rms{d}} - \Delta\tau$, and we shift the reference state to
\begin{equation}
\label{eqn:local_symmetry_map}
\Matrix{M}_k(\Delta\tau)
=
\Matrix{U}(\Delta\tau)\<
\Matrix{N}_k\<
\Matrix{\Phi}_k(\Delta t_{\rms{d}})\<
\Inv{\Matrix{\Phi}}_k(\Delta t_{\rms{d}} - \Delta\tau).
\end{equation}
The transformation $\Matrix{M}_k(\Delta\tau)$ remains isochronous but need not equal $\Matrix{N}(t_k+\Delta\tau)$, since the two factors of $\Matrix{\Phi}_k$ do not in general compose into $\Matrix{\Phi}_k(\Delta\tau)$.
Substituting \Cref{eqn:local_symmetry_map} into \Cref{eqn:local_trajectory} and evaluating at $\Delta t_{\rms{d}} - \Delta\tau$ reproduces the measurement, since $\Matrix{U}(\Delta t_{\rms{d}} - \Delta\tau)\<\Matrix{U}(\Delta\tau) = \Matrix{U}(\Delta t_{\rms{d}})$ and the factors of $\Matrix{\Phi}_k$ cancel.
Varying $\Delta\tau$ traces a one-dimensional family of delay--state pairs that produce the same measurement.
This family is an exact symmetry of the measurement model, so a single measurement cannot separate the delay from the state.
The biases enter through $\Matrix{\Phi}_k$ and are also not identifiable from a single measurement.

Measurements at distinct times can break the symmetry, unless a single shift in the navigation state compensates for the delay change at every delayed time.
A lower bound on the number of measurements needed to determine all $16$ coordinates follows from a counting argument.
The perturbation vector has $16$ entries: nine for $\Matrix{N}_k$, one for the delay, and six for the biases.
A full-state measurement supplies nine residuals, a pose measurement six, and a position-only measurement three, so at least two, three, and six measurements are needed, respectively.
These counts are necessary conditions only, and the trajectory must also be sufficiently informative over the measurement interval.

\subsection{Measurement Nullspace Structure}
\label{subsec:nullspace_structure}

The symmetry appears to first order as a null direction of the measurement Jacobian.
We perturb the navigation state, the delay, and the biases using the coordinates of \Cref{eqn:perturbation}.
The full-state measurement Jacobian with respect to these 16 coordinates has nine rows; see~\cite{2026_Kelly_Identifiability} for the full expressions.
To isolate the delay--state symmetry, we set $\delta\Vector{b}_{\omega} = \delta\Vector{b}_{\rms{a}} = \Zero$.
Differentiating \Cref{eqn:local_symmetry_map} at $\Delta\tau = 0$, with $\dot{\Matrix{U}}(0) = \Matrix{\Gamma}$ and
\begin{equation*}
\left.
\frac{d}{d\<\Delta\tau}
\Inv{\Matrix{\Phi}}_k(\Delta t_{\rms{d}} - \Delta\tau)
\right|_{\Delta\tau = 0}
=
\Matrix{\Omega}(t_{\rms{d}})\<
\Inv{\Matrix{\Phi}}_k(\Delta t_{\rms{d}}),
\end{equation*}
and left-multiplying by $\Inv{\Matrix{N}}_k$ to obtain a right perturbation, gives
\begin{equation}
\label{eqn:null_direction}
\Inv{\Matrix{N}}_k\<\Matrix{\Gamma}\<\Matrix{N}_k
+
\Matrix{\Phi}_k(\Delta t_{\rms{d}})\<
\Matrix{\Omega}(t_{\rms{d}})\<
\Inv{\Matrix{\Phi}}_k(\Delta t_{\rms{d}})
=
\bbm \Vector{\mu} \\ 0 \ebm^{\wedge}.
\end{equation}
The time components of the two terms in \Cref{eqn:null_direction} add to zero, so $\Vector{\mu} \in \Real^9$.
The measurement is unchanged to first order when $\delta\Vector{\zeta}_k = \Vector{\mu}\<\delta\tau$, which keeps the perturbed transformation in the isochronous subgroup.
The delay--state null direction of the reduced Jacobian is therefore spanned by $(\Vector{\mu},\,1)$ in the coordinates $(\delta\Vector{\zeta}_k,\,\delta\tau)$.

For several measurements, local observability requires the stacked measurement Jacobian to have full column rank.
The direction $\Vector{\mu}$ depends on the delayed time through $\Matrix{\Phi}_k(\Delta t_{\rms{d}})$ and $\Matrix{\Omega}(t_{\rms{d}})$, so measurements at distinct delayed times generally contribute distinct null directions, and the stacked Jacobian generically attains full column rank once the counts of \Cref{subsec:measurement_symmetry} are met.%
\footnote{When the inputs are constant, every measurement shares the same null direction, and the system is unobservable regardless of the number of measurements.}
The active window serves this purpose, retaining enough of the trajectory to relate several delayed measurements to the same transformations.

The delayed times must also fall within the window.
Aiding measurements available at update $k$ constrain only the portion of the window ending $\tau$ before the current time, so the usable interval is the window duration minus $\tau$.
The window duration should therefore exceed the largest anticipated delay by enough margin to retain the required number of well-separated measurements.

\begin{figure}[t!]
\centering
\includegraphics[width=\columnwidth,trim={130px, 58px, 56px, 96px},clip]%
{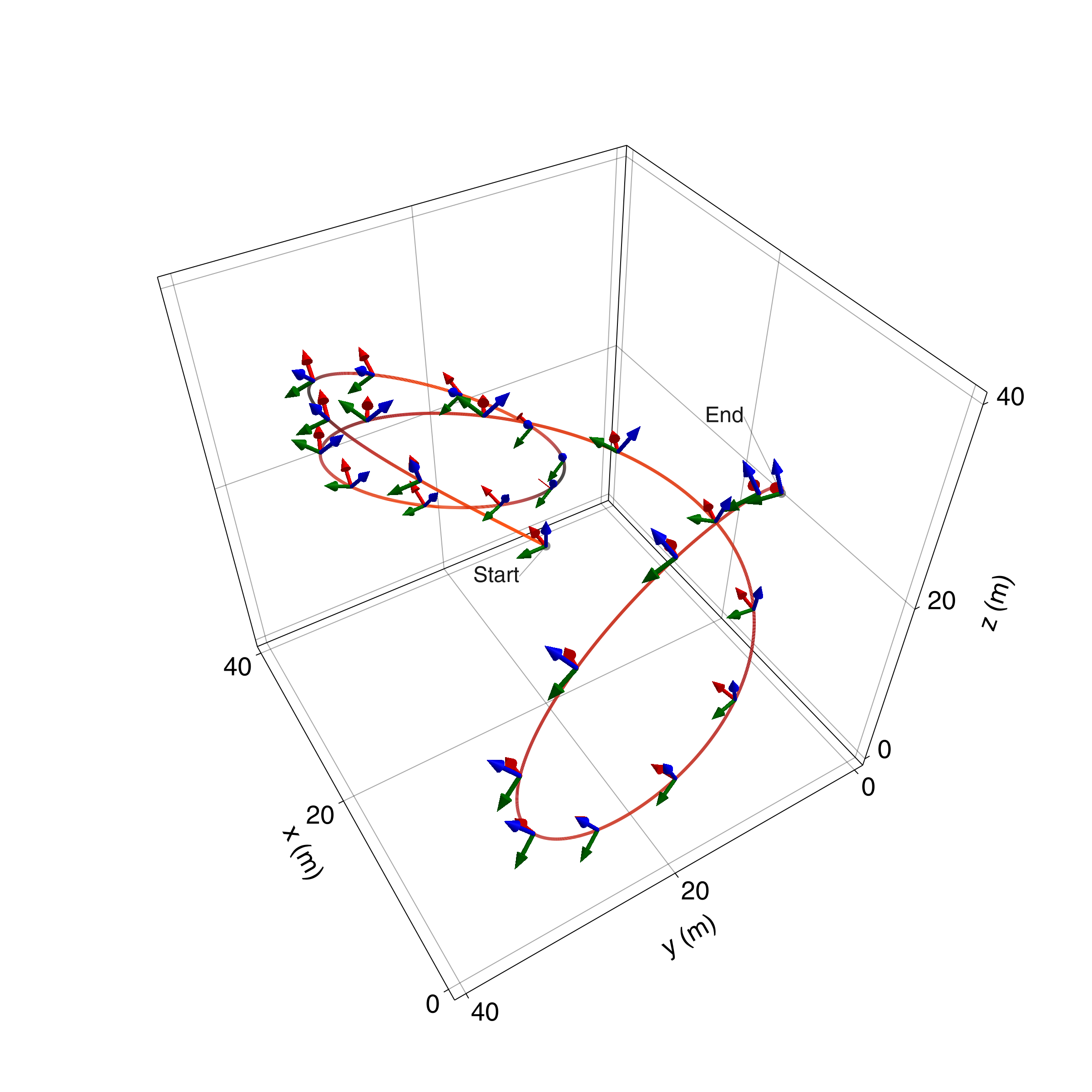}
\caption{The \emph{Smooth} trajectory used in the simulation studies.
The \emph{Excited} and \emph{Aggressive} trajectories (not shown) superimpose additional accelerations on this trajectory.
Body-frame axes are shown with $x$ forward (red), $y$ right (green), and $z$ up (blue); the curve is shaded by the magnitude of the velocity relative to the fixed inertial frame.}
\label{fig:trajectory}
\vspace*{-4mm}
\end{figure}

\begin{figure*}[t!]
\label{fig:performance_plots}
\centering
\includegraphics[width=0.98\textwidth]{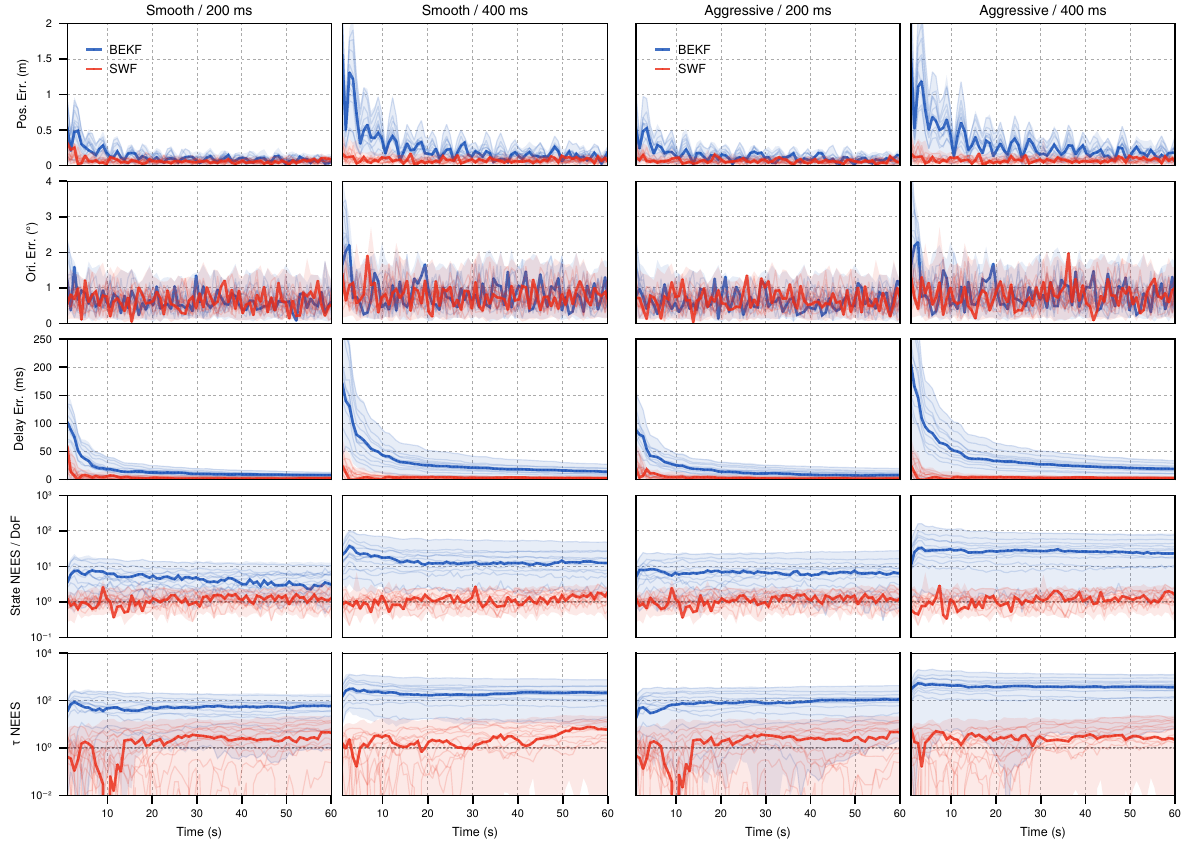}%
\vspace*{-1mm}%
\caption{Monte Carlo comparison of the buffered EKF (BEKF) and the proposed sliding window filter (SWF) for delayed aided inertial navigation.
Columns show the \emph{Smooth} and \emph{Aggressive} trajectories with delays of $200$ ms and $400$ ms.
Rows show absolute position error, orientation error, delay error, state NEES/DoF, and $\tau$ NEES over time.
Blue denotes the BEKF and red the SWF.
Shaded bands indicate the range over all trials, faint curves show a subset of individual trials, and solid curves show representative trials; only every eighth time sample is plotted for clarity.}
\vspace{-3mm}
\end{figure*}

\section{Simulation Studies}
\label{sec:simulation_studies}

We use simulation to evaluate the accuracy and consistency of the proposed filter, with particular attention to the measurement update structure required for observability.
We compare the sliding window filter (SWF) with the buffered EKF (BEKF) described in \Cref{sec:related_work}.
The BEKF navigation state, propagation, and delayed-state reconstruction respect the appropriate Galilean group structure.
A vanilla augmented-state EKF is not evaluated, since our prior work established its poorer performance on this problem~\cite{2021_Kelly_Question}.

\subsection{Simulation Setup}
\label{subsec:simulation_setup}

The simulations use three trajectories and measurement delays of $200$ ms and $400$ ms.
The \emph{Smooth} trajectory, shown in \Cref{fig:trajectory}, has a low-frequency oscillating profile.
The \emph{Excited} trajectory superimposes additional sinusoidal motion on the \emph{Smooth} trajectory, increasing the trajectory curvature in some regions, while the \emph{Aggressive} trajectory introduces stronger acceleration and deceleration along the nominal path.
These trajectories provide different types of excitation relevant to delay observability~\cite{2026_Kelly_Identifiability}.

The SWF retains five transformations separated by $0.3$ s, spanning $1.2$ s, and jointly applies delayed aiding corrections across the active window.
Updates are enabled once at least six aiding measurements lie within the active window.

For each trajectory--delay--filter combination, we run 100 trials over $60$ s.
The initial position, velocity, and orientation errors are drawn from independent zero-mean Gaussian distributions with per-axis standard deviations of $1$ m, $0.1$ m/s, and $0.1$ rad, respectively.
The delay estimate is initialized at zero with a standard deviation of $0.2$ s, while the gyroscope- and accelerometer-bias errors have standard deviations of $0.01$ rad/s and $0.05$ m/s$^2$, respectively.
For each trial, both filters receive identical initial perturbations, measurements, and sensor-noise realizations.

The simulated IMU operates at $200$ Hz, and the aiding sensor provides position, velocity, and orientation measurements at $10$ Hz with per-axis standard deviations of $0.10$ m, $0.10$ m/s, and $0.01$ rad, respectively.
Each aiding measurement is additionally subject to independent timing jitter with a standard deviation of $5$ ms.

Performance is assessed using trajectory-averaged RMS position, velocity, and orientation errors and the final delay RMSE.
Consistency is assessed using the average normalized estimation error squared (ANEES) per degree of freedom for the latest 16-dimensional substate, the delay ANEES, and the delay $3\<\sigma$ coverage, defined as the fraction of estimates for which the true delay lies within three standard deviations of the estimate.

\subsection{Results and Discussion}
\label{subsec:results_discussion}

The Monte Carlo results are summarized in \Cref{tab:monte_carlo_trials}, with representative time histories shown in \Cref{fig:performance_plots}.
Three BEKF trials for the \emph{Excited} trajectory with a $400$ ms delay produce delay estimates outside the finite IMU-buffer support and are excluded from the aggregate statistics.
Over the remaining trials, the SWF generally achieves lower errors, with position RMS of $0.072$--$0.086$ m and delay RMSE of $3.22$--$3.41$ ms, compared with $0.092$--$0.456$ m and $3.70$--$20.24$ ms for the BEKF.

The consistency results show a much larger difference.
For the latest 16-dimensional navigation--delay--bias substate, the SWF has ANEES/DoF between $1.21$ and $1.40$, whereas the BEKF ranges from $1.99$ to $31.53$.
For the delay alone, the SWF has $\tau$ ANEES of $3.50$--$5.89$ and $3\<\sigma$ coverage of $76.7$--$89.1\%$, compared with $12.97$--$472.81$ and $0.2$--$46.1\%$ for the BEKF.
The SWF is therefore still somewhat overconfident in the delay estimate, but avoids the severe spurious information gain exhibited by the recursive BEKF.

Although the BEKF becomes highly inconsistent, its navigation errors remain bounded in the retained trials.
As seen in \Cref{fig:performance_plots}, the global aiding measurements continue to correct position, velocity, and orientation, while process noise prevents complete covariance contraction.
The BEKF can therefore maintain bounded navigation errors despite a biased delay estimate and substantial overconfidence.

The results also show that more aggressive motion does not necessarily provide more information about the delay.
The \emph{Excited} trajectory improves BEKF delay estimation relative to both the \emph{Smooth} and \emph{Aggressive} trajectories, consistent with the trajectory-dependent observability analysis of \Cref{sec:observability} and~\cite{2026_Kelly_Identifiability}.
In contrast, the additional along-path acceleration and deceleration of the \emph{Aggressive} trajectory provide little additional delay information relative to the \emph{Smooth} trajectory.
Greater trajectory informativeness improves BEKF accuracy but does not fix its consistency problem.
The key distinction is how the estimators use past trajectory information: the BEKF processes reconstructed delayed states recursively, whereas the SWF preserves cross-time correlations and incorporates multiple delayed measurements jointly before marginalization, substantially reducing spurious information gain.

\begin{table*}[t]
\centering
\caption{Monte Carlo performance for the buffered EKF (BEKF) and sliding window filter (SWF).
One hundred trials were run for each trajectory--delay--filter combination.
Position, velocity, and orientation errors are RMS values over the full trajectory; delay RMSE is evaluated at the final time.
State ANEES/DoF is computed using the latest 16-dimensional substate, while delay ANEES and $3\<\sigma$ coverage are averaged over the trajectory and trials.
The $\dagger$ entry excludes three BEKF trials in which the estimated delay left the finite IMU-buffer support; all other entries use all 100 trials.}
\label{tab:monte_carlo_trials}
\vspace{2mm}
\renewcommand{\arraystretch}{1.02}
\setlength{\tabcolsep}{4pt}
\centering
\begin{tabular}{@{}cccccccccc@{}}
\toprule
Trajectory &
Delay (ms) &
Filter &
Pos.\ RMS (m) &
Vel.\ RMS (m/s) &
Ori.\ RMS ($^\circ$) &
$\tau$ RMSE (ms) &
State ANEES/DoF &
$\tau$ ANEES &
$\tau$ $3\<\sigma$ Cov.\ (\%) \\
\midrule
\multirow{4}{*}{Smooth}
& \multirow{2}{*}{200} & BEKF         & 0.175 & 0.175 & 0.719 & 7.93 & 4.97 & 60.43 & 9.0 \\
&                          & SWF          & 0.072 & 0.101 & 0.684 & 3.22 & 1.21 & 3.50 & 89.1 \\
\cmidrule(lr){2-10}
& \multirow{2}{*}{400} & BEKF         & 0.404 & 0.362 & 0.967 & 14.50 & 17.19 & 245.92 & 0.2 \\
&                          & SWF          & 0.086 & 0.120 & 0.828 & 3.36 & 1.23 & 3.92 & 87.2 \\
\midrule
\multirow{4}{*}{Excited}
& \multirow{2}{*}{200} & BEKF         & 0.092 & 0.102 & 0.702 & 3.70 & 1.99 & 12.97 & 46.1 \\
&                          & SWF          & 0.072 & 0.106 & 0.682 & 3.37 & 1.40 & 5.79 & 77.9 \\
\cmidrule(lr){2-10}
& \multirow{2}{*}{400} & BEKF$^\dagger$ & 0.182 & 0.162 & 0.882 & 5.07 & 3.68 & 32.01 & 18.9 \\
&                          & SWF          & 0.086 & 0.121 & 0.825 & 3.41 & 1.39 & 5.89 & 76.7 \\
\midrule
\multirow{4}{*}{Aggressive}
& \multirow{2}{*}{200} & BEKF         & 0.192 & 0.186 & 0.721 & 10.21 & 7.76 & 104.93 & 9.4 \\
&                          & SWF          & 0.072 & 0.101 & 0.684 & 3.22 & 1.21 & 3.51 & 89.1 \\
\cmidrule(lr){2-10}
& \multirow{2}{*}{400} & BEKF         & 0.456 & 0.402 & 0.987 & 20.24 & 31.53 & 472.81 & 1.6 \\
&                          & SWF          & 0.086 & 0.120 & 0.828 & 3.36 & 1.23 & 3.93 & 87.1 \\
\bottomrule
\end{tabular}
\vspace{-3.5mm}
\end{table*}

\section{Conclusion}
\label{sec:conclusion}

We presented a sliding window filter on $\LieGroupSGal{3}$ for aided inertial navigation with an unknown constant measurement delay.
The filter retains a short history of navigation states and uses interpolation to apply delayed aiding corrections jointly across the active window.
This construction is motivated by our observability analysis, which shows that, for a single delayed measurement, the model admits an exact symmetry: a change in the delay can be compensated by a change in the navigation state, leaving the measurement unchanged.
Measurements at multiple times can provide sufficiently independent constraints to eliminate this unobservable direction.
The window retains these cross-time constraints explicitly while keeping complexity bounded through marginalization, albeit at greater computational cost than standard recursive filters.

Our simulation studies reveal substantial differences in estimator consistency.
The buffered EKF is able to reconstruct past transformations for delayed updates, but it can become highly inconsistent.
The resulting severe overconfidence demonstrates that reconstructing past transformations from a finite IMU history is not, by itself, enough to fix the underlying observability issue.
By comparison, the sliding window markedly improves consistency while maintaining accurate delay and state estimates.

There are several directions for future work.
It would be valuable to examine how window length and trajectory excitation affect observability.
Other extensions include alternative interpolation methods and joint calibration of the delay and an unknown sensor extrinsic transformation.
Evaluation on a physical platform is also an important next step, particularly to assess robustness to real-world timing jitter, time-varying latency, and other modelling errors present in practical sensing systems.

\printbibliography

\end{document}

%% file: notation-shortcuts.tex
\usepackage{xparse}

\let\originalleft\left
\let\originalright\right
\renewcommand{\left}{\mathopen{}\mathclose\bgroup\originalleft}
\renewcommand{\right}{\aftergroup\egroup\originalright}

\NewDocumentCommand\Real{}{ \mathbb{R} }

\NewDocumentCommand\bbm{}{ \begin{bmatrix} } 
\NewDocumentCommand\ebm{}{ \end{bmatrix} }   
\NewDocumentCommand\T{}{\mathsf{T}}          

\NewDocumentCommand\Vector{m}{ \boldsymbol{\mathbf{#1}} }

\NewDocumentCommand\Matrix{m}{ \bm{\mathbf{#1}} }

\NewDocumentCommand\Inv{m}{{#1}^{-1}}

\NewDocumentCommand\Zero{}{ \Matrix{0} }
\NewDocumentCommand\Identity{}{ \Matrix{I} }

\NewDocumentCommand\LieGroupSO{m}{ \mathrm{SO}(#1) }
\NewDocumentCommand\LieAlgebraSO{m}{ \mathfrak{so}(#1) }

\NewDocumentCommand\LieGroupSE{m}{ \mathrm{SE}(#1) }

\NewDocumentCommand\LieGroupSETwo{m}{ \mathrm{SE_{2}}(#1) }

\NewDocumentCommand\LieGroupSGal{m}{ \mathrm{SGal}(#1) }
\NewDocumentCommand\LieAlgebraSGal{m}{ \mathfrak{sgal}(#1) }

\NewDocumentCommand\Matlog{m}{\mathrm{ln}\left(#1\right)}

\NewDocumentCommand\Matexp{m}{\exp\left(#1\right)}

\NewDocumentCommand\Defined{}{ \triangleq }

%% file: notation-specifics.tex
\usepackage{scalerel}

\NewDocumentCommand\<{}{\mspace{1mu}}

\newcommand{\rms}[1]{\mathrm{#1}}

\NewDocumentCommand\State{}{ \Vector{\mathcal{X}} }

\NewDocumentCommand\SmallState{}{\scaleobj{0.85}{\State}}
\NewDocumentCommand\Perturbed{m}{\tilde{#1}}